\pdfoutput=1
\documentclass[11pt]{article}

\usepackage[final]{acl}

\usepackage{times}
\usepackage{latexsym}

\usepackage[T1]{fontenc}
\usepackage[utf8]{inputenc}

\usepackage{microtype}

\usepackage{inconsolata}

\usepackage{graphicx}

\usepackage{hyperref}
\usepackage{url}
\usepackage{booktabs}       % professional-quality tables
\usepackage{amsfonts}       % blackboard math symbols
\usepackage{nicefrac}       % compact symbols for 1/2, etc.
\usepackage{lipsum}
\usepackage{fancyhdr}       % header
\usepackage{float}
\usepackage{multirow}
\usepackage{placeins}
\usepackage{enumitem}
\usepackage{arydshln}       % for dashed lines
\usepackage{verbatim}
\usepackage{listings}
\usepackage{xcolor}
\usepackage{caption}
\usepackage{wrapfig}
\usepackage[most]{tcolorbox}

\hypersetup{
    colorlinks=true, % 设置为true表示链接带颜色,设为false则表示链接以框的形式呈现
    linkcolor=blue, % 内部链接（例如目录到章节的链接）的颜色设置为蓝色
    filecolor=magenta, % 文件链接的颜色
    urlcolor=cyan, % 外部URL链接的颜色
    citecolor=blue % 引用链接的颜色
}

\newcommand{\ie}{\emph{i.e., }}
\newcommand{\eg}{\emph{e.g., }}

\newcommand{\cf}{\emph{cf. }}

\newtcolorbox{prompt}[1]{
    enhanced,
    colback=gray!20,
    colframe=black,
    boxrule=0.3pt,
    arc=3mm,
    left=2pt,
    right=2pt,
    boxsep=3pt,
    fonttitle=\small\bfseries,
    title=#1,
    fontupper=\scriptsize
}

\title{Route Sparse Autoencoder to Interpret Large Language Models}

\author{
 \textbf{Wei Shi\textsuperscript{1}}\thanks{Equal Contribution},
 \textbf{Sihang Li\textsuperscript{1}}\footnotemark[1],
 \textbf{Tao Liang\textsuperscript{2}},
 \textbf{Mingyang Wan\textsuperscript{2}},
\\
 \textbf{Guojun Ma\textsuperscript{2}}\thanks{Corresponding},
 \textbf{Xiang Wang\textsuperscript{1}}\footnotemark[2],
 \textbf{Xiangnan He\textsuperscript{1}},
\\
\\
 \textsuperscript{1}University of Science and Technology of China,
 \textsuperscript{2}Douyin Co., Ltd.,
\\
\texttt{swei2001@mail.ustc.edu.cn}, 
\texttt{taoliangdpg@126.com}, \\
\texttt{\{wanmingyang, maguojun\}@bytedance.com},\\
\texttt{\{sihang0520, xiangwang1223, xiangnanhe\}@gmail.com} 
}

\begin{document}
\maketitle

\begin{abstract}
Mechanistic interpretability of large language models (LLMs) aims to uncover the internal processes of information propagation and reasoning. 
Sparse autoencoders (SAEs) have demonstrated promise in this domain by extracting interpretable and monosemantic features. 
However, prior works primarily focus on feature extraction from a single layer, failing to effectively capture activations that span multiple layers.
In this paper, we introduce Route Sparse Autoencoder (RouteSAE), a new framework that integrates a routing mechanism with a shared SAE to efficiently extract features from multiple layers. 
It dynamically assigns weights to activations from different layers, incurring minimal parameter overhead while achieving high interpretability and flexibility for targeted feature manipulation.
We evaluate RouteSAE through extensive experiments on Llama-3.2-1B-Instruct. 
Specifically, under the same sparsity constraint of 64, RouteSAE extracts 22.5\% more features than baseline SAEs while achieving a 22.3\% higher interpretability score. 
These results underscore the potential of RouteSAE as a scalable and effective method for LLM interpretability, with applications in feature discovery and model intervention.
Our codes are available at \url{https://github.com/swei2001/RouteSAEs}.
\end{abstract}
\section{Introduction}
Mechanistic interpretability of large language models (LLMs) seeks to understand and intervene in the internal process of information propagation and reasoning, to further improve trust and safety \cite{toyModels, neurons, gpt2IOI}.
Sparse autoencoders (SAEs) identify causally relevant and interpretable monosemantic features in LLMs, offering a promising solution for mechanistic interpretability \cite{claudeTowards}. 
Therefore, SAE and its variants \cite{vanillaSAE, gatedSAE, topkSAE, jumpreluSAE} have been widely utilized in LLM interpretation tasks, such as feature discovery \cite{claudeScaling, topkSAE} and circuit analysis \cite{saeCircuits}.

Typically, SAE is trained in an unsupervised manner.
It first disentangles the intermediate activations from a single layer in the language model into a sparse, high-dimensional feature space, which is subsequently reconstructed by a decoder. 
This process reverses the effects of superposition \cite{superposition} by extracting features that are sparse, linear, and decomposable.
\begin{figure}[t]
    \centering
    \includegraphics[width=0.47\textwidth]{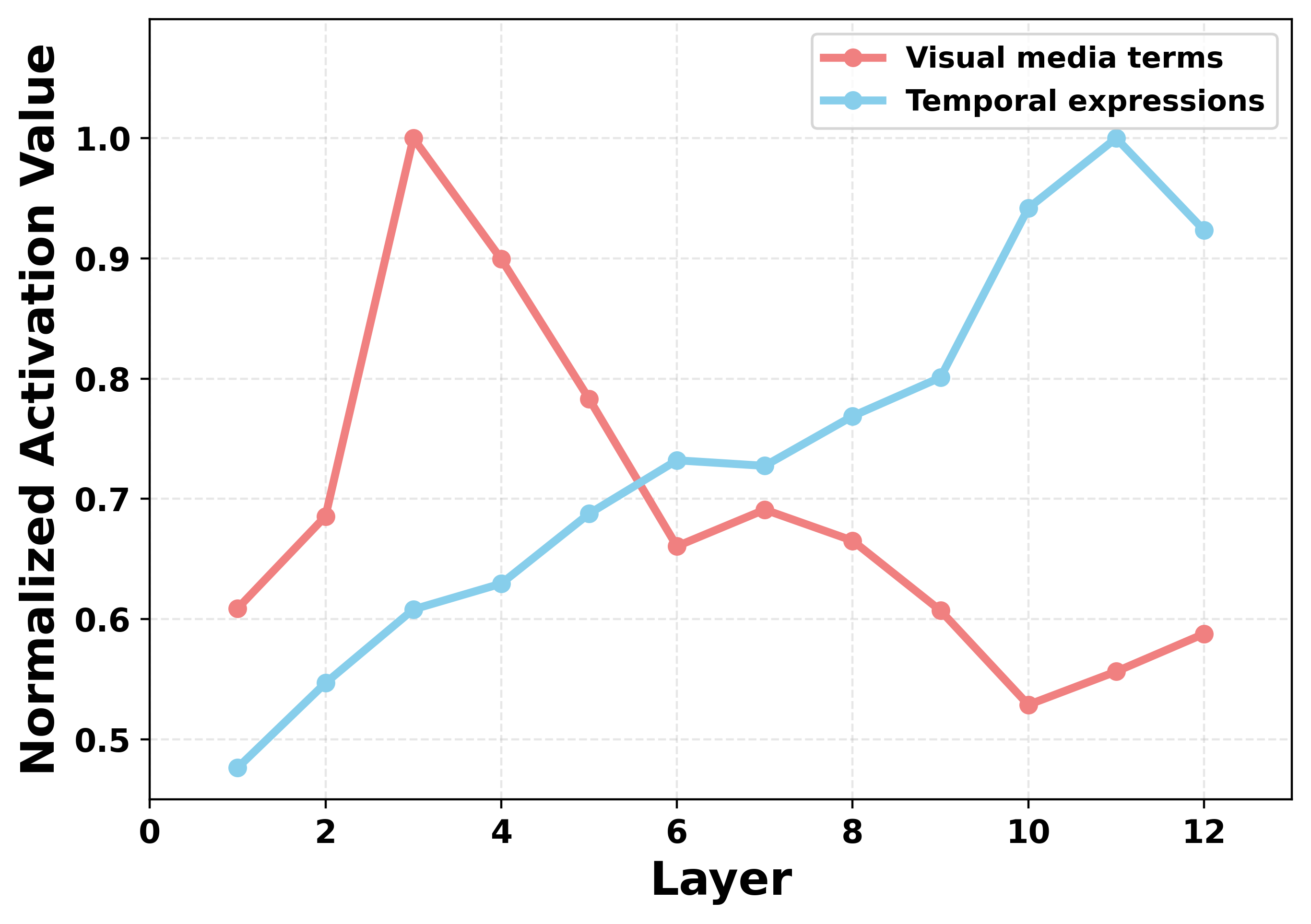}
    \vspace{-9pt}
    \caption{Layer-wise normalized activation values for two features extracted by Topk SAE in \texttt{pythia-160m}. The low-level feature (visual media terms) exhibits high activation in early layers that gradually decreases in deeper layers. 
    In contrast, the high-level feature (temporal expressions) shows increasing activation with depth, peaking in the later layers.}
    \label{fig:teaser}
\end{figure}

\begin{figure*}
    \centering
    \includegraphics[width=0.97\linewidth]{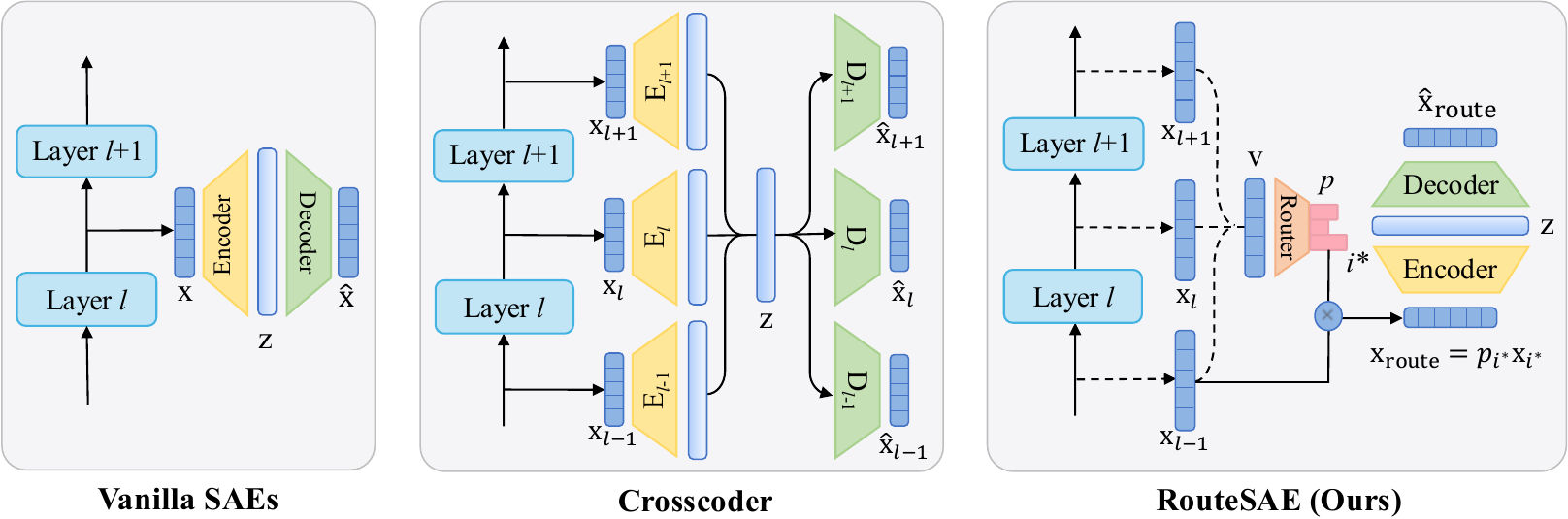}
    \vspace{-9pt}
    \caption{Comparison of vanilla single-layer SAE, Crosscoder, and RouteSAE.
    Most existing SAEs belong to the vanilla SAE category, where features are extracted from the activation of a single layer.
    Crosscoder relies on separate encoders and decoders for each layer.
    RouteSAE incorporates a lightweight router to dynamically integrate multi-layer residual stream activations.}
    \label{fig:comparison}
\end{figure*}
However, the activation strength of features in this feature space exhibits distinct distribution patterns across layers\footnote{Referred to as ``Transformer factors'' in \cite{transformerVis}.} \cite{transformerVis}. 
As shown in Figure \ref{fig:teaser}, low-level features, which are associated with disambiguating word-level polysemy, tend to exhibit peak activation in the early layers and decline steadily in deeper layers.
High-level features, which capture sentence-level or long-range structure, show increasing activation with depth.\footnote{Refer to \cite{transformerVis} for more examples of low- and high-level features.}

This distribution disparity presents a significant challenge for previous SAEs \cite{vanillaSAE, gatedSAE, topkSAE, jumpreluSAE}, as they typically extract features from the hidden state of a single layer, failing to capture feature activating at other layers effectively (\cf Figure \ref{fig:comparison}).
Recently proposed Sparse Crosscoders\footnote{Currently a conceptual framework without complete experimental validation.} \cite{crosscoder} serve as an alternative to address this limitation, which separately encodes the hidden states of each layer into a high-dimensional feature space and aggregates the resulting representations for reconstruction (\cf Figure \ref{fig:comparison}). 
This approach facilitates the joint learning of features across different layers.
However,  Crosscoder has two critical limitations: 
(1) \textbf{Limited scalability}: For an $L$-layer model, Crosscoder employs $L$ separate encoders and decoders to process activations layer by layer, resulting in a parameter scale approximately $L$ times larger than traditional SAEs. 
This significantly increases computational overhead during both training and inference. 
(2) \textbf{Uncontrollable interventions}: Crosscoder's joint learning mechanism projects hidden states into a high-dimensional space and then aggregates them, making it impractical to precisely identify and adjust the feature activations at specific layers. This limits its flexibility for tasks requiring controlled, feature-level interventions, \eg feature steering~\cite{claudeScaling}.

To address these challenges, we propose \textbf{Route Sparse Autoencoder (RouteSAE)}.
At the core is integrating a lightweight router with a shared SAE to dynamically extract multi-layer features in an efficient and flexible manner.
A router is employed to compute normalized weights for activations from multiple layers.
This dynamic weighting approach significantly reduces the number of parameters compared to a suite of layer-specific encoders and decoders, thereby addressing scalability concerns.
Additionally, by unifying feature disentanglement and reconstruction within a shared SAE, RouteSAE facilitates fine-grained adjustments of specific feature activations, enabling more controlled interventions to influence the model's output.
This enhances flexibility and supports precise feature-level control, making the framework well-suited for tasks requiring robust and interpretable manipulation of model activations.

We conduct comprehensive experiments on Llama-3.2-1B-Instruct \cite{llama3}, evaluating downstream KL divergence, interpretable feature numbers, and interpretation score.
The experimental results demonstrate that RouteSAE significantly improves the interpretability.
At an equivalent sparsity level of 64, it achieves a 22.5\% increase in the number of interpretable features and a 22.3\% improvement in interpretation scores.
% Additionally, we present a case study to highlight the practical advantages of the routing mechanism, illustrating how it improves feature interpretability by capturing features across multiple layers.

% 第七段：总结我们的贡献
Our contributions are summarized as follows:
\begin{itemize}[leftmargin=*]
    \item We propose RouteSAE, a novel sparse autoencoder framework that integrates multi-layer activations through a routing mechanism.
    \item RouteSAE achieves higher computational efficiency than Crosscoder by using a shared SAE structure with minimal additional parameters.
    \item Extensive experiments confirm that RouteSAE enhances model interpretability, highlighting the effectiveness of the proposed routing mechanism.
\end{itemize}

\section{Methodology}
\begin{figure*}
    \centering
    \includegraphics[width=0.95\linewidth]{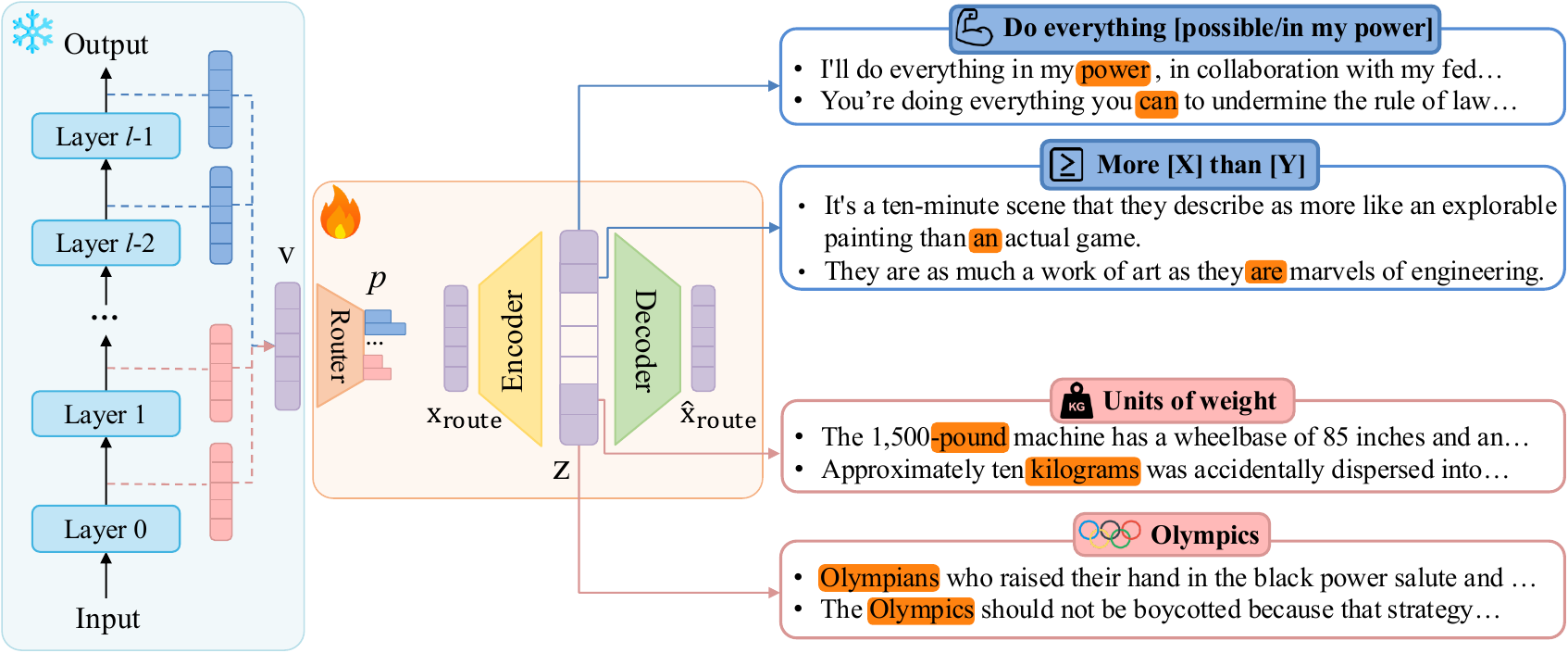}
    \vspace{-9pt}
    \caption{RouteSAE employs a lightweight router to dynamically integrate activations from multiple residual stream layers, effectively disentangling them into a shared feature space. 
    It enables the model to capture features across different layers --- low-level features such as ``units of weight'' and ``Olympics'' from shallow layers, and high-level features like ``more [X] than [Y]'' and ``do everything [possible/in my power]'' from deeper layers.}
    \label{fig:framework}
\end{figure*}

In this section, we first briefly review SAEs, then introduce our proposed Route Sparse Autoencoder (\textbf{RouteSAE}) in detail.

\subsection{Preliminary}
\label{sec:prelim}

\textbf{SAE and Feature Decomposition.} 
SAEs decompose language model activations --- typically residual streams \cite{resnet}, $\mathbf{x} \in \mathbb{R}^d$, into a sparse linear combination of features $\mathbf{f}_1, \mathbf{f}_2, \ldots, \mathbf{f}_M \in \mathbb{R}^d$, where $M \gg d$ represents the feature space dimension. 
The original activation $\mathbf{x}$ is reconstructed using an encoder-decoder pair defined as follows:
\begin{equation}
    \mathbf{z}=\sigma(\mathbf{W}_{\text{enc}} (\mathbf{x} - \mathbf{b}_{\text{pre}}))
\end{equation}
\begin{equation}
    \hat{\mathbf{x}}=\mathbf{W}_{\text{dec}} \mathbf{z} + \mathbf{b}_{\text{pre}},
\end{equation}
where $\mathbf{W}_{\text{enc}} \in \mathbb{R}^{M \times d}$ and $\mathbf{W}_{\text{dec}} \in \mathbb{R}^{d \times M}$ are the encoder and decoder weight matrices, $\mathbf{b}_{\text{pre}} \in \mathbb{R}^d$ is a bias term, and $\sigma$ denotes the activation function. 
The latent representation $\mathbf{z} \in \mathbb{R}^M$ encodes the activation strength of each feature. 
The training objective is to minimize the reconstruction mean squared error (MSE):
\begin{equation}
\mathcal{L} = \|\mathbf{x} - \hat{\mathbf{x}}\|_2^2.
\end{equation}

\textbf{TopK SAE.}
Early SAEs \cite{vanillaSAE, claudeTowards} leverage the ReLU activation function \cite{relu} to generate sparse feature representations, coupled with an additional $L_1$ regularization term on latent representation $\mathbf{z}$ to enforce sparsity.
However, this approach is prone to feature shrinkage, where the $L_1$ constraint drives positive activations in $\mathbf{z}$ toward zero, reducing the expressive capacity of the sparse feature space.
To mitigate this issue, TopK SAE \cite{topkSAE} replaces the ReLU activation function with a $\text{TopK}(\cdot)$ function, which directly controls the number of active latent dimensions by selecting the top $K$ largest values in $\mathbf{z}$.
This is defined as:
\begin{equation}
\mathbf{z} = \text{TopK}(\mathbf{W}_{\text{enc}} (\mathbf{x} - \mathbf{b}_{\text{pre}})).
\end{equation}
By eliminating the need for an $L_1$ regularization term, TopK SAE achieves a more effective balance between sparsity and reconstruction quality, while enhancing the model's ability to learn disentangled and interpretable monosementic features.
In our RouteSAE framework, the shared SAE module is instantiated as a TopK SAE due to its superior performance in producing monosemantic features.

\subsection{Route Sparse Autoencoder (RouteSAE)}
As shown in Figure \ref{fig:comparison}, existing SAEs are typically trained on intermediate activations from a single layer, restricting their ability to simultaneously capture both low-level features from shallow layers and high-level features from deep layers. 
To overcome this limitation, RouteSAE incorporates a lightweight router to dynamically integrate multi-layer residual streams from language models and disentangle them into a unified feature space.

\textbf{Layer Weights.}
As illustrated in Figure \ref{fig:framework}, the router receives residual streams from multiple layers and determines which layer’s activation to route. 
Instead of concatenating these activations, which could result in an excessively large input dimension, we adopt a simple yet effective aggregation strategy: sum pooling.
Specifically, given activations $\mathbf{x}_i \in \mathbb{R}^d$ from layer $i$, we aggregate them using sum pooling to form the router's input:
\begin{equation}
\mathbf{v} = \sum_{i=0}^{L-1} \mathbf{x}_i, \quad \mathbf{x}_i \in \mathbb{R}^d,
\end{equation}
where $L$ denotes the total number of layers being routed.
The resulting vector $\mathbf{v} \in \mathbb{R}^d$ serves as a condensed representation of multi-layer activations.
Next, the router projects $\mathbf{v}$ into $\mathbb{R}^L$ using a learnable weight matrix $\mathbf{W}_{\text{router}} \in \mathbb{R}^{L \times d}$, yielding the layer weight vector $\boldsymbol{\alpha}$:
\begin{equation}
\boldsymbol{\alpha} = \mathbf{W}_{\text{router}} \mathbf{v} \in \mathbb{R}^L.
\end{equation}
Each element $\alpha_i$ in $\boldsymbol{\alpha}$ represents the unnormalized weight for layer $i$, indicating its relative importance in the routing process.
These weights are then normalized using a softmax function to obtain layer selection probabilities $p_i$:
\begin{equation}
p_i = \frac{\exp(\alpha_i)}{\sum_{j=0}^{L-1} \exp(\alpha_j)}, \quad i = 0, 1, \ldots, L-1.
\label{equ:pi}
\end{equation}
$p_i$ reflects the likelihood that the activation strength peaks at layer $i$, dynamically assigned by the router based on the input representations.

\textbf{Routing Mechanisms.}
In RouteSAE, the router selects the layer $i^*$ with the highest probability $p_i$, computed as described in Equation \ref{equ:pi}.
Formally, this is expressed as:
\begin{equation}
i^* = \arg\max_{i} p_i, \quad i = 0, 1, \ldots, L-1.
\end{equation}
To ensure differentiability, we scale the activation $\mathbf{x}_{i^*}$ from the selected layer $i^*$ by its corresponding probability $p_{i^*}$, using it as input to the shared SAE for disentangling into the high-dimensional feature space and subsequent reconstruction training:
\begin{equation}
\mathbf{x}_{\text{route}} = p_{i^*} \mathbf{x}_{i^*}.
\label{equ:x-route}
\end{equation}
The latent representation $\mathbf{z}$ and the reconstruction $\hat{\mathbf{x}}$ are calculated as follows:
\begin{eqnarray}
\mathbf{z}_{\text{route}} &=& \text{TopK}(\mathbf{W}_{\text{enc}} (\mathbf{x}_{\text{route}} - \mathbf{b}_{\text{pre}}))\\
\hat{\mathbf{x}}_{\text{route}} &=& \mathbf{W}_{\text{dec}} \mathbf{z}_{\text{route}} + \mathbf{b}_{\text{pre}}.
\end{eqnarray}
Finally, we minimize the reconstruction MSE:
\begin{equation}
\mathcal{L} = \|\mathbf{x}_{\text{route}} - \hat{\mathbf{x}}_{\text{route}}\|_2^2.
\end{equation}
This objective function jointly trains the router and the shared TopK SAE, ensuring efficient and adaptive feature extraction across multiple layers.

% \textbf{Comparison of Routing Mechanisms.}
% Hard routing enforces a sparse selection by selecting only a single layer's activation, potentially focusing on the one with the strongest activation for a given input.
% In contrast, soft routing integrates information from all layers, based on their respective significance probabilities.
% Meanwhile, soft routing imposes stricter requirements on the router. 
% While hard routing only requires the router to select the layer with the highest feature activation, soft routing demands an accurate estimation of the importance of all layers.
% If the router fails to make precise predictions, it may assign disproportionately high weights to layers with low-level features strength, thus accumulating irrelevant activations and potentially misleading the subsequent disentangling of monosemantic features.

\textbf{Shared SAE and Unified Feature Space.}
The routed intermediate activation ( $\mathbf{x}_\text{route}$, as defined in Equation \ref{equ:x-route}) is processed by a shared SAE for reconstruction, which in this work is instantiated as a TopK SAE \cite{topkSAE}. 
Notably, RouteSAE is flexible and can be easily adapted to various SAE variants.
% including ReLU SAE \cite{vanillaSAE, claudeTowards}, Gated SAE \cite{gatedSAE} and JumpReLU SAE \cite{jumpreluSAE}.
By employing a shared SAE, RouteSAE establishes a unified feature space across activations from all routing layers. 
This ensures consistent feature representations, thereby enhancing the disentanglement of high-dimensional features and improving interpretability.
\section{Experiments}
We first outline the experimental setup, followed by the evaluation of RouteSAE.
In this paper, we follow prior work \cite{topkSAE, gatedSAE, vanillaSAE, claudeScaling, llamaScope} and employ multiple evaluation metrics to assess the effectiveness of RouteSAE, including downstream KL-divergence, interpretable features, interpretation score, and reconstruction loss. 
Finally, we provide a detailed case study, demonstrating that RouteSAE not only effectively captures both low-level features from shallow layers and high-level features from deep layers, but also enables targeted manipulation of these features to control the model’s output.

\subsection{Setup}
% \begin{table}[t]
%     \centering
%     \begin{tabular}{cc}
%         \toprule
%         \textbf{Model} & \textbf{Llama-3.2-1B-Instruct} \\ 
%         \midrule
%         \textbf{Hidden Size} & 2,048  \\ 
%         \textbf{\# Layers} & 16  \\ 
%         \textbf{Routing Layers} & [3:11] \\
%         \textbf{SAE Width} & 16,384 (8x) \\ 
%         \textbf{Batch Size} & 64 \\
%         \bottomrule
%     \end{tabular}
%     \vspace{-6pt}
%     \caption{Implementation details of RouteSAEs for Llama-3.2-1B-Instruct. Note that the layer indices start from 0.}
%     \vspace{-6pt}
%     \label{tab:llm_model}
% \end{table}
\textbf{Inputs.} 
We train all SAEs on the residual streams of the Llama-3.2-1B-Instruct. 
For baseline SAEs, we follow the standard approach \cite{topkSAE} of selecting the layer located approximately at $\frac{3}{4}$ of the model depth (\ie Layer 11).
Prior work \cite{robustLLMs} has shown that the early layers of LLMs primarily handle detokenization, whereas later layers specialize in next-token prediction. 
Based on this insight, we select residual streams from the middle layers of the model as input for both RouteSAE and Crosscoder \cite{crosscoder}. 
In particular, we focus on layers spanning $\frac{1}{4}$ to $\frac{3}{4}$ of the model depth, as detailed in Table \ref{tab:llm_model}.

The training data is sourced from OpenWebText2 \cite{pile}, comprising approximately 100 million randomly sampled tokens for training, with an additional 10 million tokens reserved for evaluation.
All experiments are conducted using a context length of 512 tokens. 
To ensure stable training, we normalize the language model activations following \cite{topkSAE}.

\textbf{Hyperparameters.} 
For all SAEs, we use the Adam optimizer \cite{adam} with standard settings: $\beta_1 = 0.9$ and $\beta_2 = 0.999$.
The learning rate is set to $5 \times 10^{-4}$, following a three-phase schedule.
(1) Linear warmup.
The learning rate increases linearly from 0 to the target rate over the first 5\% of training steps.
(2) Stable phase.
The learning rate remains constant for 75\% of the training steps.
(3) Linear Decay.
The learning rate gradually decreases to zero over the final 20\% of training steps to ensure smooth convergence.
To improve training stability, we apply unit norm regularization \cite{topkSAE} to the columns of the SAE decoder every 10 steps, ensuring that the decoder columns maintain unit length.

\textbf{Baselines.}
We benchmark RouteSAE against leading baselines, including ReLU SAE \cite{vanillaSAE}, Gated SAE \cite{gatedSAE}, TopK SAE, and Crosscoder \cite{crosscoder}.
Moreover, we compare with a random setting, where the router is replaced by a uniform distribution that assigns equal routing weights to each layer.
It is important to note that Crosscoder remains a conceptual framework and lacks complete experimental validation. 
As there is no official codebase or hyperparameter guidance available, we implement it following the description in \cite{crosscoder}. 
We acknowledge that our results \textit{may not fully reflect its actual performance}.
\begin{figure}[t]
    \centering
    \includegraphics[width=0.95\linewidth]{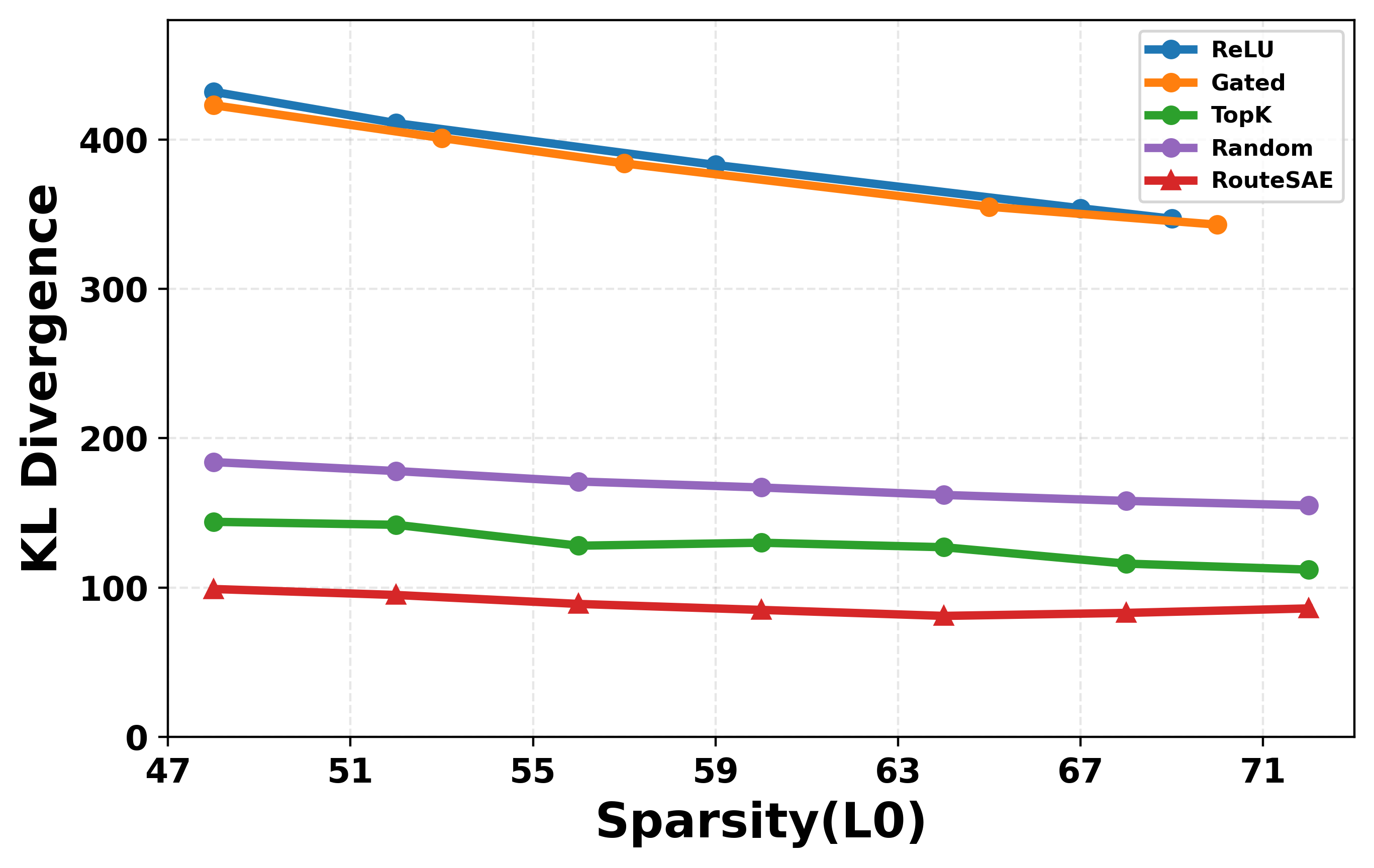}
    \vspace{-9pt}
    \caption{Pareto frontier of sparsity versus KL divergence.
    RouteSAE achieves a lower KL divergence at the same sparsity level.}
    \label{fig:1B_KLDiv}
\end{figure}

\subsection{Downstream KL Divergence}
\label{exp:downstream}
To assess whether the extracted features are relevant for language modeling, we replace the residual streams $\mathbf{x}$ with the reconstructed representation $\hat{\mathbf{x}}$ during the forward pass of the language model and evaluate the reconstruction quality using Kullback-Leibler (KL) divergence. 
It quantifies the discrepancy between the original and reconstructed distributions, with lower value indicating that the extracted features are highly relevant for language modeling.
Note that RouteSAE replaces the activation at the layer with the highest routing weight.

As shown in Figure \ref{fig:1B_KLDiv}, the sparsity-KL divergence frontiers for ReLU and Gated SAE are nearly identical, yet both exhibit a significant gap compared to TopK SAE. 
Due to suboptimal reconstruction quality, the KL divergence for ReLU and Gated SAE drops substantially as $L_0$ increases, falling from around 400 to 350. 
In contrast, the KL divergence for both TopK and RouteSAE remains consistently below 150, with only minimal decreases as $L_0$ increases.
This indicates that both methods are able to effectively reconstruct the original input $\mathbf{x}$ even at high sparsity levels.
The random routing baseline yields higher KL divergence than both TopK and RouteSAE, further highlighting the advantage of learned routing.

Notably, RouteSAE achieves the best performance among all methods, maintaining a lower KL divergence at comparable sparsity levels. 
It outperforms even TopK SAE, indicating that feature substitution during inference is most effective when performed at the layer where the target feature is most active, rather than at a predetermined fixed layer. 
We exclude Crosscoder from this comparison, as it produces multiple reconstructed representations $\hat{\mathbf{x}}$, making its application to this setting nontrivial and not directly comparable.

\subsection{Interpretable Features}
\label{exp:features}
\begin{figure}[t]
    \centering
    \includegraphics[width=0.48\textwidth]{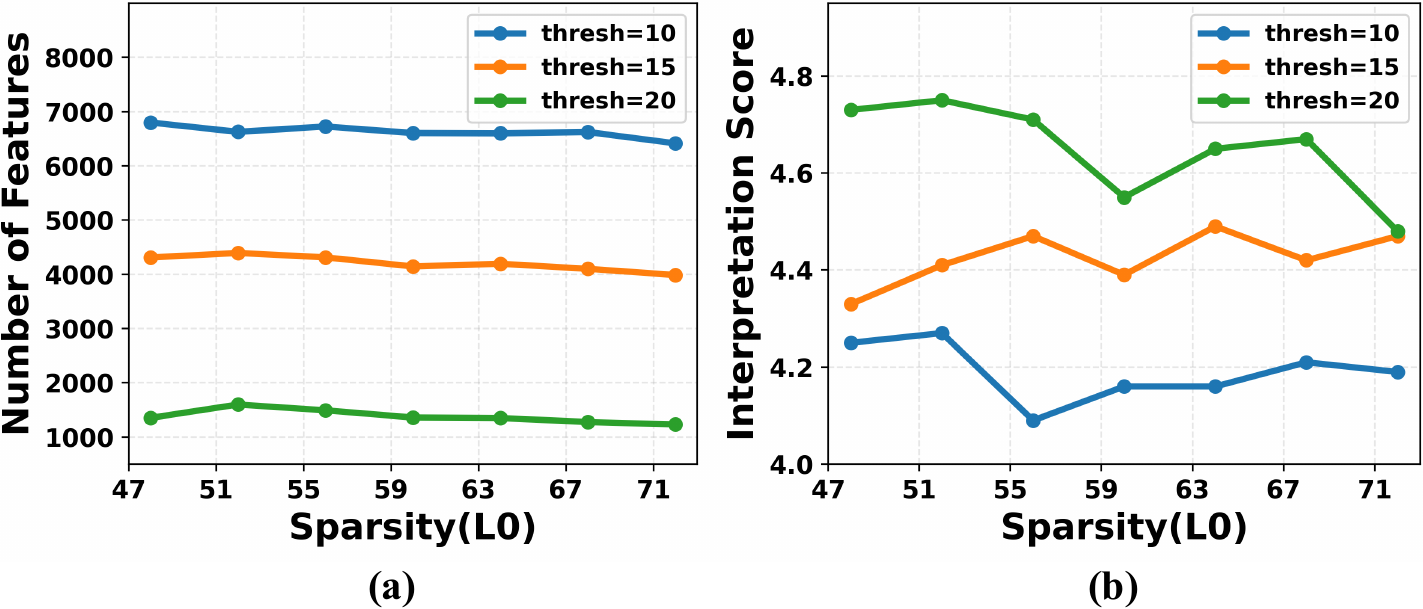}
    \caption{Effect of threshold on feature interpretability in RouteSAE. (a) Increasing the threshold reduces the number of selected features. (b) Higher thresholds yield better interpretation scores across sparsity levels.}
    \label{fig:thresh}
    \vspace{-6pt}
\end{figure}

Previous works \cite{vanillaSAE, llamaScope} interpret features by preserving the context with the highest feature activation value. 
However, we argue that it has two limitations:
(1) Retaining only the highest activation context for each feature leads to a large number of undiscernible features; 
(2) Each feature is associated with only a single context, reducing the reliability of the interpretation.

To address these limitations, we introduce a new approach for preserving feature contexts using an activation threshold. 
For a given sequence context, only features with activation values exceeding the threshold are retained.
As shown in Figure \ref{fig:thresh}(a), increasing the threshold reduces the number of retained features. 
In contrast, Figure \ref{fig:thresh}(b) demonstrates that a higher threshold leads to improved interpretation scores. 
Consequently, the threshold governs a trade-off between the quantity of interpretable features and their interpretability quality.
In this section, we set the threshold to 15, which achieves a balance between maintaining sufficient feature quantity and enhancing interpretability.
Notably, a single sequence may be associated with multiple contexts.

To further refine the interpretation, activated contexts are categorized based on their activation tokens, maintaining a min-heap of activation values.
We retain the top 2 contexts with the highest activation values within each activated token.
A filtering step is applied to remove features with fewer than four active contexts, ensuring that only sufficiently represented features are considered.
To evaluate feature extraction, we use 10 million tokens from the evaluation set to extract contexts associated with each feature.

\begin{figure}[t]
    \centering
    \includegraphics[width=.95\linewidth]{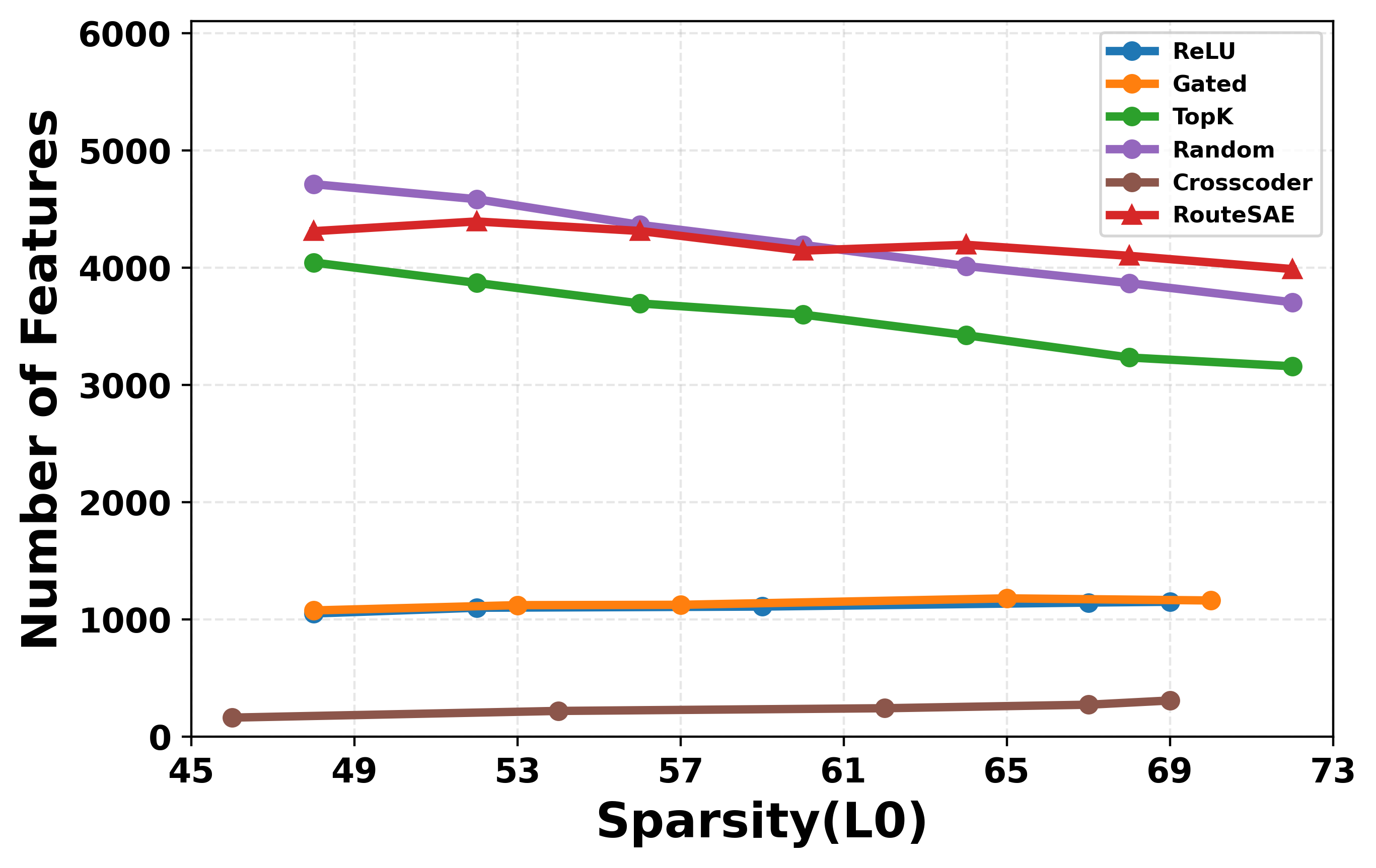}
    \vspace{-9pt}
    \caption{Comparison of the interpretable feature number.
    RouteSAE extracts the most interpretable features at the same threshold.
    }
    \vspace{-4pt}
    \label{fig:1B_features}
\end{figure}

As illustrated in Figure \ref{fig:1B_features}, at a threshold of 15, both ReLU and Gated SAE extract over 1,000 interpretable features, performing similarly.
In contrast, TopK SAE significantly outperforms both, extracting more than 3,000 features. 
RouteSAE surpasses all other methods, extracting over 4,000 features at the same threshold. 
Notably, RouteSAE exhibits a more gradual decline in the number of extracted features as $L_0$ increases, while TopK SAE exhibits a more pronounced reduction. 
The random routing baseline sometimes extracts even more features than RouteSAE, but its feature count decreases much more rapidly as $L_0$ increases.
These results suggest that learning based solely on single-layer activation values limits the ability of SAEs to extract interpretable features. 
In comparison, Crosscoder extracts substantially fewer features, retaining approximately 200. 
Since Crosscoder aggregates and projects activations across multiple layers, we hypothesize that the optimal threshold for balancing feature quantity and interpretability lies in a lower range for Crosscoder.
Therefore, comparing it against the same activation threshold may not reflect its actual ability to extract high-quality features.
We plan to investigate this in future work.

\subsection{Interpretation Score}
\label{exp:interpret_score}
\begin{figure}[t]
    \centering
    \includegraphics[width=.95\linewidth]{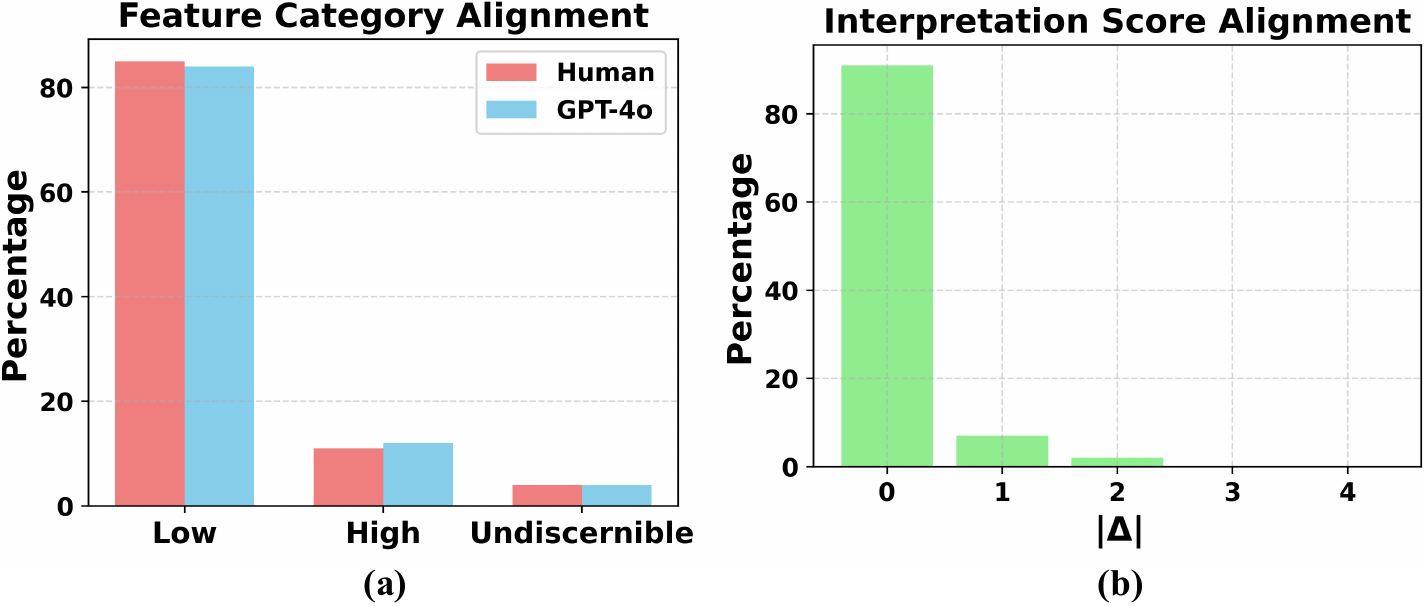}
    \vspace{-6pt}
    \caption{Human–GPT-4o alignment in the automatic feature interpretation pipeline. (a) Percentage of features assigned to each category (Low, High, or Undiscernible) by humans and GPT-4o. (b) Distribution of the absolute differences in interpretability scores between human annotators and GPT-4o.}
    \vspace{-6pt}
    \label{fig:alignment}
\end{figure}
Despite the feature screening in Section \ref{exp:features}, the number of retained features remains in the thousands, making manual interpretation and evaluation challenging. 
To further assess feature interpretability, we follow prior work \cite{vanillaSAE, claudeScaling, llamaScope} and leverage GPT-4o \cite{gpt4o} to analyze the features extracted by SAEs, assigning an interpretability score alongside feature descriptions. 
Unlike previous approaches, we provide GPT-4o with multiple token categories per feature along with their contextual usage.
Given resource constraints, we randomly select a subset of 100 retained features per SAE for interpretation. 
As detailed in Appendix \ref{app:prompt_design}, for each feature, we construct a structured prompt comprising a prefix prompt, the activated token, and its surrounding context, which is then given to GPT-4o.
GPT-4o outputs three standardized components:
(1) Feature categorization, labeling each feature as low-level, high-level, or undiscernible;
(2) Interpretability score, rated on a scale of 1 to 5; and
(3) Explanation, providing a brief justification for the assigned category and score.

To evaluate the consistency between GPT-4o and human annotators in both feature categorization and interpretability scoring, we randomly sample 100 features from RouteSAE.
For each feature, we provide its activation contexts and a scoring prompt to both GPT-4o and human annotators.
As illustrated in Figure \ref{fig:alignment}, (a) shows the percentage of features assigned to each interpretability category (``Low,'' ``High,'' or ``Undiscernible'') by both humans and GPT-4o.
The two distributions are nearly identical, reflecting strong categorical agreement between human and GPT-4o annotations.
(b) depicts the distribution of absolute differences $|\Delta|$ in interpretability scores, showing that most features exhibit minimal discrepancy between human and GPT-4o ($|\Delta| < 2$).
This indicates a high degree of alignment in interpretability assessment.

To quantify overall interpretability, we compute the average interpretability score across the 100 sampled features for each SAE. 
Due to stochasticity in both feature selection and GPT-4o’s scoring, these results should be viewed as indicative rather than definitive measures of interpretability.

\begin{figure}[t]
    \centering
    \includegraphics[width=.95\linewidth]{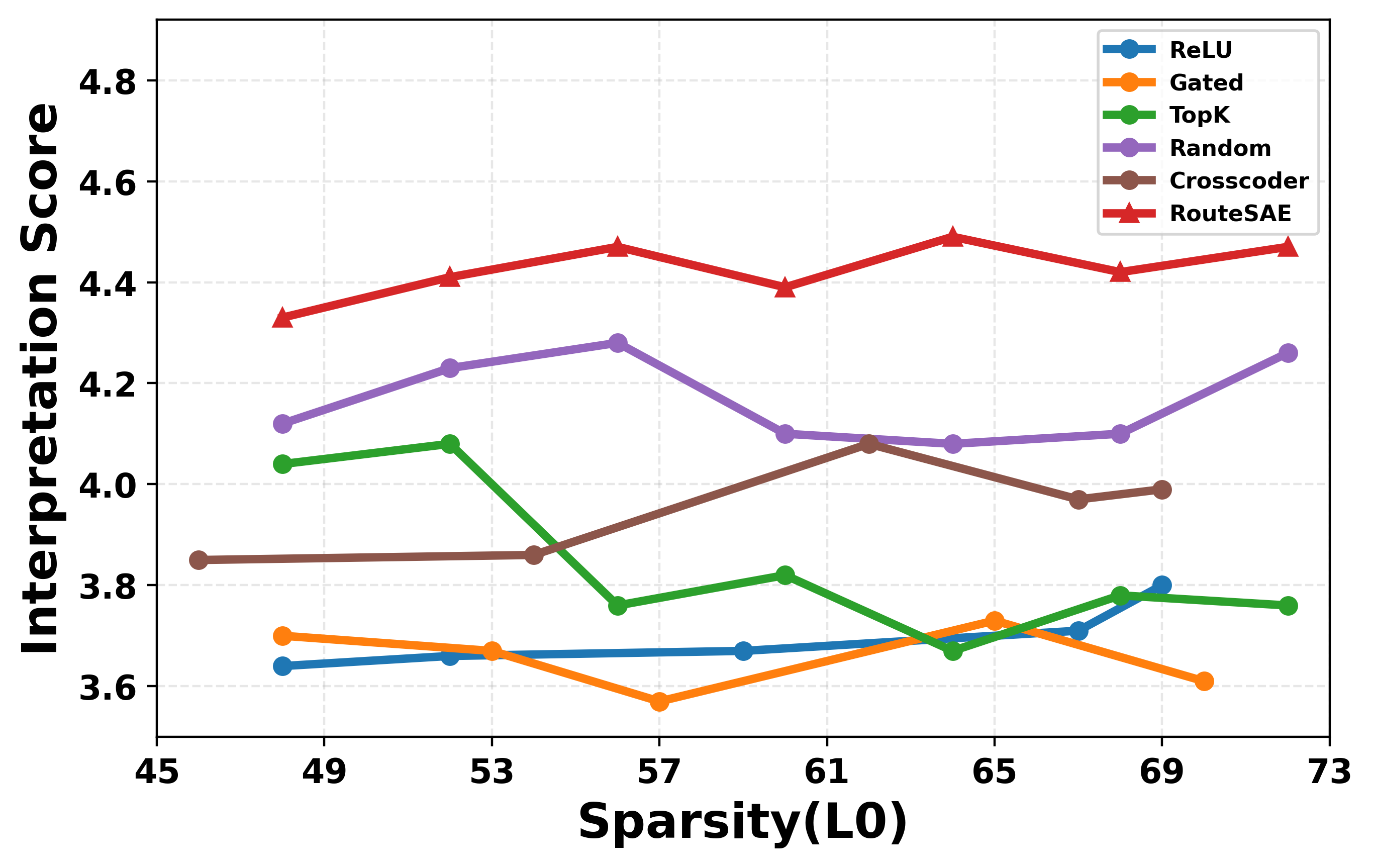}
    \vspace{-9pt}
    \caption{Comparison of interpretation scores. RouteSAE achieves a higher interpretation score at the same sparsity level.}
    \label{fig:1B_Score}
\end{figure}

Figure \ref{fig:1B_Score} shows that both ReLU and Gated SAE exhibit low and relatively stable interpretation scores, consistently falling below those of the other methods.
TopK SAE shows a noticeable decline in interpretation scores as $L_0$ increases, with scores dropping from over 4.0 at sparsity 48 to around 3.7 at sparsity 72. 
In contrast, Crosscoder, despite not being sensitive to changes in sparsity, maintains consistent scores, hovering around 3.9 across all sparsity levels.
The random routing baseline achieves higher interpretation scores than ReLU, Gated, TopK, and Crosscoder, but remains consistently lower than RouteSAE.
In comparison, RouteSAE achieves the highest interpretation scores, maintaining values above 4.4 at all sparsity levels.
It remains largely unaffected by changes in sparsity, demonstrating its robust ability to preserve high interpretability, regardless of the sparsity setting.
These results indicate that dynamically leveraging multi-layer activations, as done in RouteSAE and even to some extent by the random router, not only allows for extraction of more features but also leads to higher feature interpretability.

\subsection{Routing Weights}
\label{exp:weights}
\begin{figure}[t]
    \centering
    \includegraphics[width=.97\linewidth]{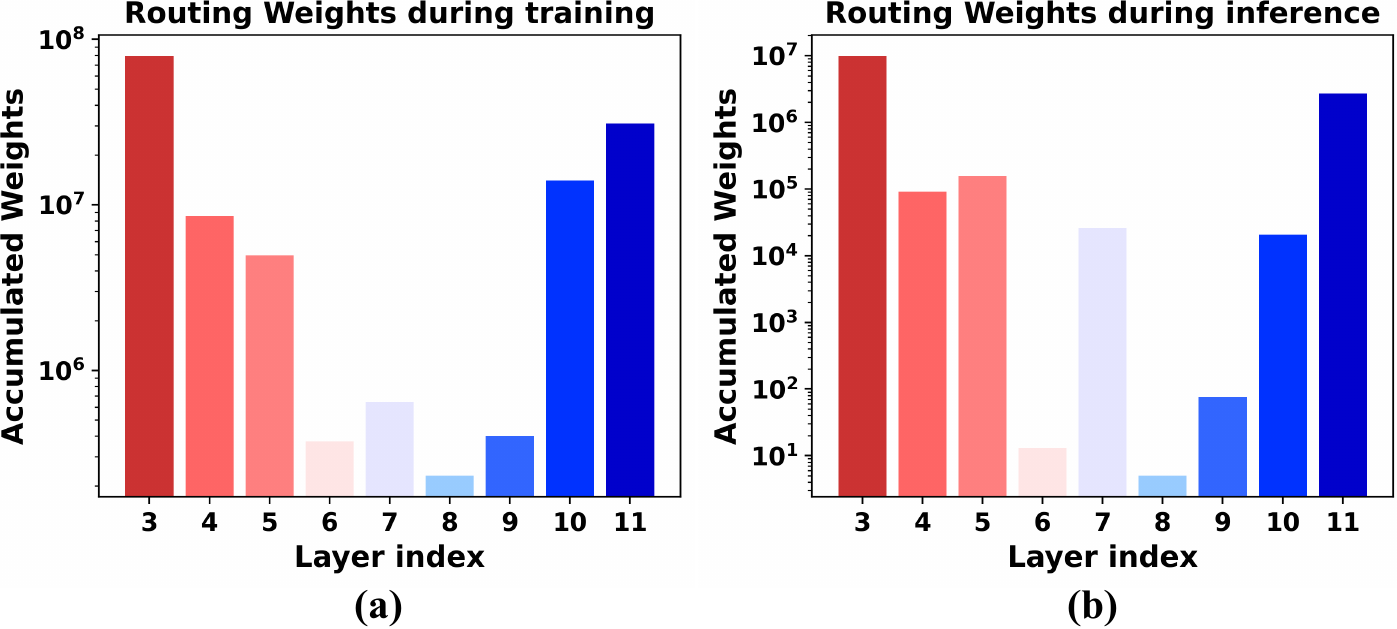}
    \caption{Illustration of weights assigned to routing layers during training (a) and inference (b). In both cases, the weights exhibit a U-shaped distribution rather than concentrating on a small subset of shallow layers.}
    \vspace{-9pt}
    \label{fig:weights}
\end{figure}

In fact, reconstructing the activations of a language model using SAE becomes increasingly difficult as the number of layers grows.
This is likely due to the increasing abstraction and entanglement of features in deeper layers, which imposes additional challenges on the autoencoder's capacity to isolate and reconstruct meaningful components.

To analyze how RouteSAE allocates routing weights across layers during training and inference, we track the layer-wise routing weights throughout both phases.
As shown in Figure \ref{fig:weights}, RouteSAE produces a distinct weight profile across layers, exhibiting a U-shaped distribution rather than concentrating weights on a small subset of shallow layers.
This pattern suggests a balanced allocation of representational capacity, where both shallow and deep layers contribute meaningfully.
These results are consistent with the observations reported in \cite{transformerVis}, which indicate that lower-level features are primarily activated in the earlier layers, whereas higher-level features become more prominent in the deeper layers.

\subsection{Case Study}
\label{exp:case}
\textbf{Interpretable Features.} As shown in Figure \ref{fig:framework}, RouteSAE effectively captures both low-level and high-level features from shallow and deep layers, respectively.
Specifically, RouteSAE identifies low-level features such as ``units of weight'' and ``Olympics'' from shallow layers. 
The ``units of weight'' feature activates on tokens related to weight units, including terms like ``pound'' and ``kilograms''. 
The ``Olympics'' feature captures variations of the term ``Olympic'', such as ``Olympics'' and ``Olympian''.
These two features exemplify word-level polysemy disambiguation, peaking at shallow layers.
At deeper layers, RouteSAE extracts high-level features, including the patterns ``more [X] than [Y]'' and ``do everything [possible/in my power]''. 
The first feature identifies tokens that appear in comparative structures, particularly those following the pattern ``more [X] than [Y].'' 
The second feature highlights tokens in phrases expressing a commitment to maximal effort or capability, such as ``do my best'', and ``do all he could''.
These two features reflect sentence-level or long-range pattern formation, peaking at deeper layers.
These observations demonstrate that RouteSAE successfully integrates features from multiple layers of activations into a unified feature space. 
For more interpretable features, refer to Appendix \ref{app:features}.

\textbf{RouteSAE Feature Steering}
\begin{figure}[t]
    \centering
    \includegraphics[width=.97\linewidth]{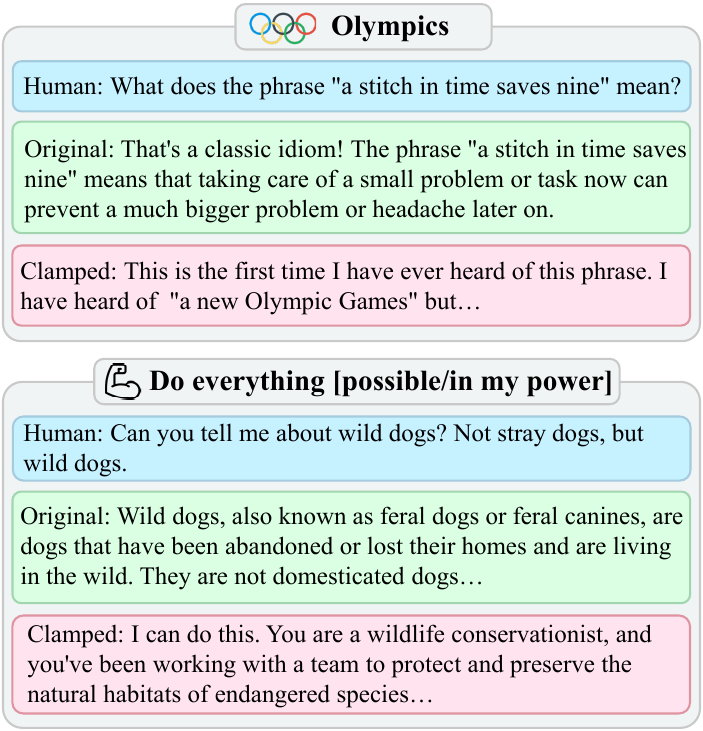}
    \caption{Illustration of feature steering via activation manipulation in RouteSAE. 
    The original response is generated with unaltered feature activations, while the clamped response is produced after setting the target feature’s activation to a high value. The upper example demonstrates a low-level feature associated with the ``Olympics” concept; increasing its activation leads the model to output Olympics-related content. The lower example involves a high-level feature linked to ``doing everything possible”; increasing its activation causes the model to adopt an \textit{all-in} attitude in its response.}
    \vspace{-9pt}
    \label{fig:steering}
\end{figure}
Figure \ref{fig:steering} illustrates how RouteSAE enables controlled model steering by directly manipulating internal features from the SAE decoder.
This is achieved by replacing the activation $\mathbf{x}$ with the reconstructed representation $\mathbf{\hat{x}}$.
In each example, the original response is generated without intervention, reflecting the model's default behavior. 
In contrast, the clamped response is obtained by increasing the activation of a specific target feature to 20. 
In the upper example, the clamped feature is a low-level one related to the ``Olympics'' concept; after intervention, the model's response becomes focused on Olympic-related content, regardless of the input question. 
In the lower example, the manipulated feature is a high-level one representing the intent to ``do everything possible''; as a result, the model adopts a proactive, determined stance, as evidenced by responses such as ``I can do this.'' 
This illustrates that, RouteSAE enables more controllable and targeted interventions on model behavior through direct feature activation manipulation.

\section{Conclusion}
In this paper, we introduce Route Sparse Autoencoder (RouteSAE), a new framework designed to enhance the mechanistic interpretability of LLMs by efficiently extracting features from multiple layers. 
Through the integration of a dynamic routing mechanism, RouteSAE enables the assignment of layer-specific weights to each routing layer, achieving a fine-grained, flexible, and scalable approach to feature extraction. 
Extensive experiments demonstrate that RouteSAE significantly outperforms traditional SAEs, with a 22.5\% increase in the number of interpretable features and a 22.3\% improvement in interpretability scores at the same sparsity level. 
These results underscore the potential of RouteSAE as a powerful tool for understanding and intervening in the internal representations of LLMs. 
By enabling more precise control over feature activations, RouteSAE facilitates better model transparency and provides a solid foundation for future work in feature discovery and interpretability-driven model interventions.

% \section*{Impact Statement}
% This paper presents work whose goal is to advance the field of Machine Learning. 
% There are many potential societal consequences of our work, none which we feel must be specifically highlighted here.

% \section*{Acknowledgments}
% This work is supported by...

\section*{Limitations}
While RouteSAE shows promising results, several limitations remain, which we aim to address in future research.

\textbf{Improvements of the router.} 
To the best of our knowledge, we are the first to introduce a routing mechanism in SAEs to learn a shared feature space. 
However, we employed a simple linear projection, which has limited capabilities. 
Our experiments show that the weight distribution of the router is influenced by the feature space size $M$ and the sparsity level $k$. 
Therefore, exploring more sophisticated activation aggregation methods and router designs is an important direction for future work.
    
\textbf{Cross-layer features.}
Research on cross-layer feature extraction is still in its early stages, and the current method of dynamically selecting activations across multiple layers, as presented in this paper, is not yet optimized for discovering cross-layer features. 
Further exploration is needed to enable RouteSAE to more effectively identify and utilize cross-layer features.

\section*{Ethical Considerations}
This paper presents work whose goal is to advance the field of Machine Learning. 
There are many potential societal consequences of our work, none which we feel must be specifically highlighted here.

\bibliography{anthology, custom}
% \bibliography{custom}
\newpage
\appendix
\section{Related Work}
In this section, we begin by reviewing prior work on sparse encoding, followed by a discussion of SAEs for interpreting LLMs. 
Finally, we briefly introduce cross-layer feature extraction in LLMs.
\subsection{Sparse Encoding}
% 参考TopK SAE文章related work章节的第一段，重点表述早期非LLM工作
Dictionary learning \cite{dictionaryLearning} is a foundational machine learning approach that aims to learn an overcomplete set of basis components, enabling efficient data representation through sparse linear combinations.
Autoencoders \cite{autoencoder}, in contrast, are designed to extract low-dimensional embeddings from high-dimensional data. 
By merging these two paradigms, sparse autoencoders have been developed, incorporating sparsity constraints such as $L_1$ regularization \cite{L1_autoencoder} to enforce sparsity in learned representations.
% 一句话总结一下SAE的用处或者应用，承上启下
Sparse autoencoders have found widespread application across various domains of machine learning, including computer vision \cite{sparse_coding_cv} and natural language processing \cite{sparse_coding_nlp}.

\subsection{Sparse Autoencoder for LLMs}
% 尽可能全面地从vanilla SAE开始介绍所有SAE的变种
SAEs have emerged as effective tools for capturing monosemantic features \cite{superposition}, making them increasingly popular in LLM applications.
Early work \cite{vanillaSAE} introduced SAEs for extracting interpretable features from the internal activations of GPT-2 \cite{gpt2}. 
To address systematic shrinkage in feature activations inherent in traditional SAEs \cite{vanillaSAE, claudeTowards}, Gated SAEs \cite{gatedSAE} were proposed, decoupling feature detection from magnitude estimation.
TopK SAEs \cite{topkSAE}, inspired by k-sparse autoencoders \cite{k-sparse}, directly controlled sparsity to enhance reconstruction fidelity while preserving sparse representations.
JumpReLU SAEs \cite{jumpreluSAE} advanced the trade-off between reconstruction quality and sparsity by replacing the conventional ReLU activation \cite{relu} with the discontinuous JumpReLU function \cite{jumprelu}.
More recently, Switch SAEs \cite{switchSAE} introduced a mixture-of-experts mechanism, where inputs are routed to smaller, specialized SAEs, achieving better reconstruction performance within fixed computational constraints.
% 一到两句话说明一下当前SAE的主要问题，扣到intro中我们提到的点：从单一层捕捉，没有考虑在不同层激活的特征
However, these approaches capture the intermediate activations of language models from a single layer, neglecting features activated across multiple layers, which limits their overall applicability.

\subsection{Features across Layers}
% 从TransVis开始强调层间特征差异，然后介绍Gemma- Llama-scope这两个采用每层都训练一个SAE的工作，然后介绍CrossCoder。
Layer-wise differences in activation features within the transformer-based language model were first highlighted in \cite{transformerVis}, revealing that shallow layers capture low-level features while deeper layers focus on high-level patterns.
Building on this, Gemma Scope \cite{gemmaScope} leveraged JumpReLU SAEs \cite{jumpreluSAE} to train separate models for each layer and sub-layer of the Gemma 2 models \cite{gemma2}.
Similarly, Llama Scope \cite{llamaScope} trained 256 SAEs per layer and sublayer of the Llama-3.1-8B-Base model \cite{llama3}, extending layer-wise sparse modeling.
% 一到两句话说明一下这样训N个SAE的坏处是啥：比如计算消耗大，或者给定一个输入时不好判断该用toolkit中哪一个SAE去解释
Nevertheless, training a suite of SAEs is computationally expensive and often learns redundant features, posing significant scalability challenges for larger models. 
Moreover, determining the specific SAE relevant to a given input or characteristic can be nontrivial, complicating their practical application.
Recently, Sparse Crosscoders \cite{crosscoder} introduced a cross-layer SAE variant designed to investigate layer interactions and shared features \cite{claudeScaling, model_diffing}.
This framework facilitates circuit-level analysis \cite{claudeCircuits, saeCircuits} by enabling feature tracking across layers, providing valuable insights into the evolution of model features and architectural differences. 
% 一句话简要说一下Crosscoder的局限性
% 一到两句话总结：正是因为这个section介绍的这些工作有xxx问题，所以we are motivated to 开发RouteSAE
However, Crosscoder still relies on separate encoders and decoders for each layer, which limits its efficiency and hinders seamless integration with downstream tasks.

The challenges of scalability, feature localization, and applicability to downstream tasks  motivate the development of RouteSAE.

\section{Comparison of Routing Mechanisms.}
\label{app:routing_comp}
\begin{figure*}[ht]
    \centering
    \includegraphics[width=0.95\linewidth]{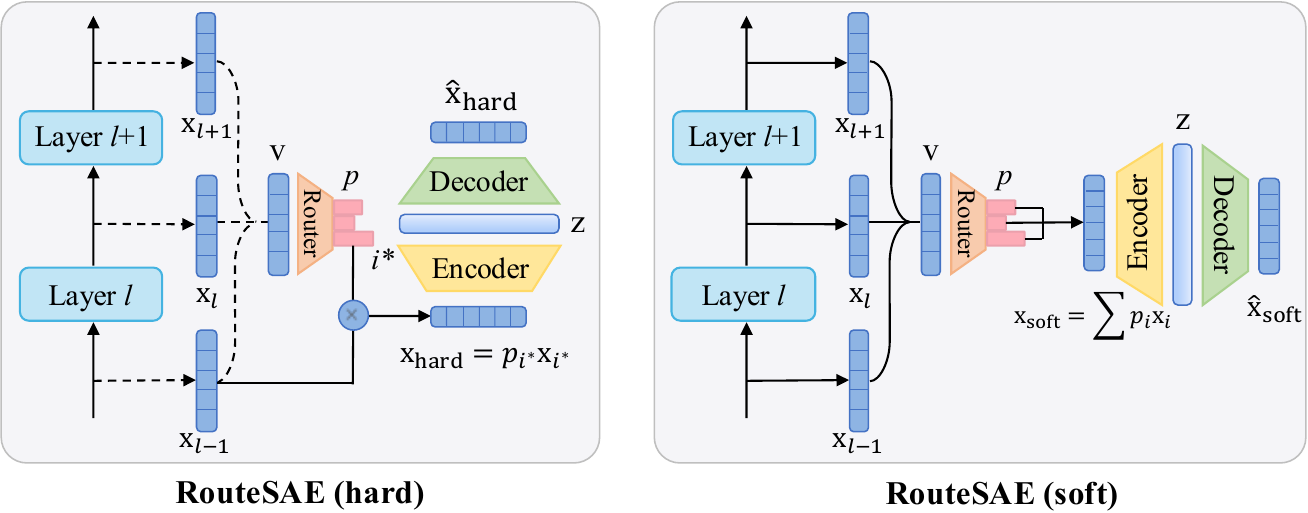}
    \vspace{-9pt}
    \caption{Routing mechanism comparison. Hard routing enforces sparse selection by activating only a single layer, whereas soft routing integrates information from all layers, weighted by their respective significance probabilities.}
    \label{fig:hard_soft}
\end{figure*}
In RouteSAE, the router determines how the multi-layer activations are integrated into the SAE. 
We denote the routing mechanism defined in Equation \ref{equ:x-route} as hard routing.

\textbf{Hard Routing.}
In hard routing, the router selects the layer with the highest probability $p_i$.
The activation $\mathbf{x}_{i^*}$ from the selected $i^*$ is scaled by its corresponding probability $p_{i^*}$ and used as the input to the SAE:
\begin{equation}
\mathbf{x}_{\text{SAE}} = p_{i^*} \mathbf{x}_{i^*}.
\end{equation}

\textbf{Soft Routing.}
As an alternative, we also explore soft routing, where the router combines activations from all layers by weighting them with their respective probabilities $p_i$. 
Instead of selecting a single layer, the input to the SAE is computed as a weighted sum of all layer activations:
\begin{equation}
\mathbf{x}_{\text{SAE}} = \sum_{i=0}^{L-1} p_i \mathbf{x}_i.
\end{equation}
This approach allows the SAE to incorporate multi-layer information in a more continuous manner, leveraging a richer feature representation compared to hard routing.

\textbf{Discussion.}
Hard routing enforces sparsity by selecting the activation from a single layer, typically the one with the strongest response for a given input. 
This mechanism simplifies the routing task, as the router only needs to identify the layer with the highest activation. 
In contrast, soft routing aggregates activations from all layers, weighted by their estimated importance scores. 
This introduces a significantly more challenging requirement: the router must accurately estimate the relative contribution of each layer. 
Inaccurate estimations may result in disproportionately high weights assigned to less relevant layers, which can lead to the accumulation of noisy or irrelevant activations. 
This, in turn, may interfere with the disentanglement of monosemantic features in subsequent stages. 
While soft routing has the potential to capture cross-layer features—i.e., features that are distributed across multiple layers—our experiments thus far have not demonstrated clear benefits in this setting. 
We plan to investigate this direction further in future work.
\begin{table}[t]
    \centering
    \begin{tabular}{cc}
        \toprule
        \textbf{Model} & \textbf{Llama-3.2-1B-Instruct} \\ 
        \midrule
        \textbf{Hidden Size} & 2,048  \\ 
        \textbf{\# Layers} & 16  \\ 
        \textbf{Routing Layers} & [3:11] \\
        \textbf{SAE Width} & 16,384 (8x) \\ 
        \textbf{Batch Size} & 64 \\
        \bottomrule
    \end{tabular}
    \vspace{-6pt}
    \caption{Implementation details of RouteSAEs for Llama-3.2-1B-Instruct. Note that the layer indices start from 0.}
    \vspace{-6pt}
    \label{tab:llm_model}
\end{table}

\section{Reconstruction Loss}
\label{app:reconstruct}
\begin{figure}[t]
    \centering
    \includegraphics[width=.95\linewidth]{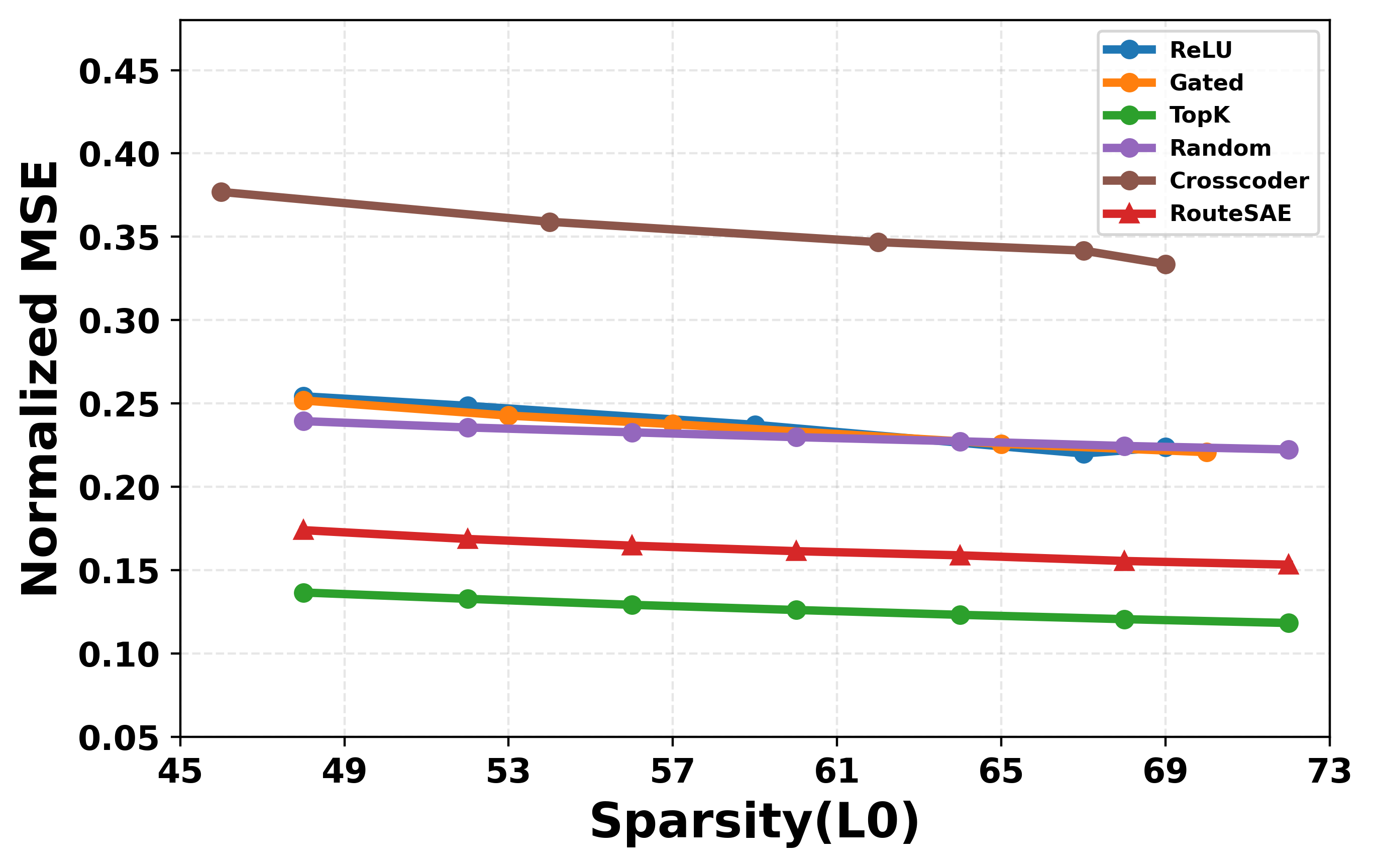}
    \vspace{-9pt}
    \caption{Pareto frontier of sparsity versus Norm MSE. Norm MSE, as a proxy metric, cannot be directly compared between models with distinct input distributions.}
    \vspace{-9pt}
    \label{fig:1B_MSE}
\end{figure}

Given a fixed sparsity $L_0$ in the latent representation $\mathbf{z}$, a lower reconstruction loss indicates better performance in terms of the SAE's ability to reconstruct the original input.
However, evaluating the effectiveness of SAEs remains challenging. 
The sparsity-reconstruction frontier is commonly used as a proxy metric, but it should be noted that the primary goal of SAEs is to extract interpretable features, not simply to reconstruct activations.
As shown in Figure \ref{fig:1B_MSE}, TopK SAE achieves the optimal sparsity-reconstruction trade-off, maintaining a normalized MSE of around 0.15 across sparsity levels. 
The performance of Random, ReLU and Gated SAE is comparable, with all three methods showing a normalized MSE of approximately 0.25, significantly lagging behind TopK. 
Crosscoder, on the other hand, demonstrates a notably poorer reconstruction frontier, with its MSE consistently around 0.35.

It is important to clarify that, as a proxy metric, normalized MSE cannot be directly compared between models with different input distributions. 
Both RouteSAE and Crosscoder receive and reconstruct activations from multiple layers, which leads to a more complex distribution compared to a single layer. 
This increased complexity makes reconstruction more difficult, resulting in a higher MSE loss.
Nevertheless, while both Crosscoder and RouteSAE aggregate activations across multiple layers, RouteSAE exhibits significantly better reconstruction performance than Crosscoder, trailing only slightly behind TopK. 
RouteSAE maintains a normalized MSE of around 0.18, demonstrating its ability to handle the complexities of multi-layer reconstruction.

\section{Auto Intrepretation Prompt Design.}
\label{app:prompt_design}

\begin{prompt}{}
\textbf{\small Background}

We are analyzing the activation levels of features in a neural network, where each feature activates certain tokens in a text.
Each token’s activation value indicates its relevance to the feature, with higher values showing stronger association. Features are categorized as:\\
A. Low-level features, which are associated with word-level polysemy disambiguation (e.g., "crushed things", "Europe").\\
B. High-level features, which are associated with long-range pattern formation (e.g., "enumeration", "one of the [number/quantifier]")\\
C. Undiscernible features, which are associated with noise or irrelevant patterns.\\
\end{prompt}

\begin{prompt}{}
\textbf{\small Task description}

Your task is to classify the feature as low-level, high-level or undiscernible and give this feature a monosemanticity score based on the following scoring rubric:\\
Activation Consistency\\
5: Clear pattern with no deviating examples\\
4: Clear pattern with one or two deviating examples\\
3: Clear overall pattern but quite a few examples not fitting that pattern\\
2: Broad consistent theme but lacking structure\\
1: No discernible pattern\\
Consider the following activations for a feature in the neural network.\\
Token: ...  Activation: ...  Context: ...\\

\textbf{\small Question}

Provide your response in the following fixed format:\\
Feature category: [Low-level/High-level/Undiscernible]\\
Score: [5/4/3/2/1]\\
Explanation: [Your brief explanation]\\
\end{prompt}

\section{Interpretable Features Extracted by RouteSAE.}
\label{app:features}
In this section, we present additional interpretable features extracted by RouteSAE from Llama-3.2-1B-Instruct, including feature-activated tokens, contexts, values, and GPT-4 explanations.

%%%%%%%%%%%%%%%%%%%%%%%%% Low-Level %%%%%%%%%%%%%%%%%%%%%%%%%
\subsection{Low-Level Features}
\label{app:low_level}
% 3675
\begin{prompt}{Feature 3675: flourish and thrive}
\textbf{\small Explanation:}
The feature consistently activates on variations of the words ``flourish'' and ``thrive'', which are semantically similar and often used interchangeably in contexts indicating growth or success. The activation values are consistently high across all instances, with no deviating examples, indicating a clear pattern associated with word-level polysemy disambiguation related to these terms.

\textbf{\small Contexts:}
Anti-Nafta rhetoric doesn't play well in El Paso, San Antonio and Houston, which have become gateway cities for commerce with Latin America and have \textbf{\textit{flourished}} since the North American Free Trade Agreement passed Congress in 1993.
\textbf{\small Activation:}
16.16

\textbf{\small Contexts:}
It's not, by the way, a song about devil-worshipping, although the Stones \textbf{\textit{thrived}} on the controversy and didn't do much to discourage speculation.
\textbf{\small Activation:}
17.33

\textbf{\small Contexts:}
When the researchers planted worn-out cattle fields in Costa Rica with a sampling of local trees, native species began to move in and \textbf{\textit{flourish}}, raising the hope that destroyed rainforests can one day be replaced.
\textbf{\small Activation:}
16.43
\end{prompt}

% 3896
\begin{prompt}{Feature 3896: academic or job application}
\textbf{\small Explanation:}
The feature consistently activates on tokens related to the context of academic or job application processes, specifically focusing on ``applicant'' and ``interviews.'' There is a clear pattern with no deviating examples, indicating a strong association with word-level polysemy disambiguation related to the application process.

\textbf{\small Contexts:}
ON a Sunday morning a few months back, I interviewed my final Harvard \textbf{\textit{applicant}} of the year.
\textbf{\small Activation:}
15.97

\textbf{\small Contexts:}
Then you have to advertise a position or opportunity, and weed through the \textbf{\textit{applicants}} to find the 5\% that are actually worth talking to.
\textbf{\small Activation:}
15.80

\textbf{\small Contexts:}
I might be smart and qualified, but for some random reason I may do poorly in the \textbf{\textit{interviews}} and not get an offer!
\textbf{\small Activation:}
15.45
\end{prompt}

% 4574
\begin{prompt}{Feature 4574: spatial or temporal prepositions}
\textbf{\small Explanation:}
The feature consistently activates on the tokens ``in'' and ``within'', indicating a strong association with spatial or temporal prepositions. The activations are highly consistent across different contexts, showing no deviating examples, which suggests a clear pattern related to the usage of these prepositions. This aligns with low-level features focused on word-level polysemy disambiguation.

\textbf{\small Contexts:}
The show was getting huge, and just as with COMDEX, the show-\textbf{\textit{within}}-a-show was born.
\textbf{\small Activation:}
17.48

\textbf{\small Contexts:}
According to a Circuit City employee in Chicago, the consumer electronics chain is trading in HD DVD players bought into their stores ``\textbf{\textit{within}} 3 months of the announcement'', as opposed to their 30-day return policy.
\textbf{\small Activation:}
28.23

\textbf{\small Contexts:}
There's now at least a 50\% risk that prices will decline \textbf{\textit{within}} two years in 11 major metro areas, including San Diego; Boston; Long Island, N.Y.; Los Angeles; and San Francisco, according to PMI Mortgage Insurance's latest U.S.
\textbf{\small Activation:}
29.30
\end{prompt}

%%%%%%%%%%%%%%%%%%%%%%%%% High-Level %%%%%%%%%%%%%%%%%%%%%%%%%
\subsection{High-Level Features}
\label{app:high_level}
% 19
\begin{prompt}{Feature 19: enumeration or distribution}
\textbf{\small Explanation:}
The feature consistently activates tokens that are part of a pattern involving enumeration or distribution, such as ``each'', ``neither'', ``all'', and ``both''. These tokens are often used in contexts where items or actions are being listed or compared, indicating a high-level feature related to long-range pattern formation. The activations show a clear pattern with no deviating examples, suggesting a strong monosemanticity.

\textbf{\small Contexts:}
A caller, discussing how Clinton and Obama are both terrifying or whatever, made the comment that “my 12-year-old says that Obama looks like Curious George!” As my jaw hit the steering wheel, Rush chuckled and they moved on to the \textbf{\textit{next}} topic.
\textbf{\small Activation:}
17.73

\textbf{\small Contexts:}
Advanced Graphics Card Repair Now that you have already learned how to repair broken capacitors and inductors on your graphics cards (or any other boards), it's time to move \textbf{\textit{on}} to the smaller components that are harder to tackle.
\textbf{\small Activation:}
16.62

\textbf{\small Contexts:}
Creating a useful command line tool Now that we have the basics out of the way, we can move \textbf{\textit{onto}} creating a tool to solve a specific problem.
\textbf{\small Activation:}
16.37
\end{prompt}

% 1424
\begin{prompt}{Feature 1424: date expressions}
\textbf{\small Explanation:}
The activations consistently highlight tokens that are part of date expressions, specifically the day of the month in a date format (e.g., ``January 1'', ``February 28'', ``March 31''). This indicates a clear pattern of recognizing and activating on numerical day components within date contexts, which aligns with high-level features associated with long-range pattern formation, such as recognizing structured data formats like dates. There are no deviating examples, hence the highest score for activation consistency.

\textbf{\small Contexts:}
As of January \textbf{\textit{1}}, more than one of every 100 adults is behind bars, about half of them Black.
\textbf{\small Activation:}
22.78

\textbf{\small Contexts:}
The Random Destructive Acts FAQ Updated March 19, 2003: It has been about 8 years since I wrote this page (before 2002 the last modification date was June \textbf{\textit{30}}, 1995) and I still get emails about it every few days.
\textbf{\small Activation:}
20.76

\textbf{\small Contexts:}
Taguba, USA (Ret.) served 34 years on active duty until his retirement on 1 \textit{}\textbf{\textit{January}} 2007.
\textbf{\small Activation:}
15.03
\end{prompt}

% 2271
\begin{prompt}{Feature 2271: comparative or equality expressions}
\textbf{\small Explanation:}
The activations consistently highlight tokens that are part of comparative or equality expressions, such as ``just as [adjective/adverb] as'' and ``equal [noun].'' This indicates a clear pattern of identifying long-range patterns related to comparisons and equality, with no deviating examples.

\textbf{\small Contexts:}
a big Obama supporter, and I would have voted the old John McCain over Hillary Clinton (but not the new, party-line-toeing, I'm-just\textbf{\textit{-as}}-conservative-as-Bush-I-swear John McCain).
\textbf{\small Activation:}
18.64

\textbf{\small Contexts:}
\textbf{\textit{Equally}} important, it represents the anticipation of how much new money will be created in the future.
\textbf{\small Activation:}
18.11

\textbf{\small Contexts:}
It was important to us to have an equal \textbf{\textit{amount}} of diversity in the cast.
\textbf{\small Activation:}
16.23
\end{prompt}

\end{document}